\documentclass{article}

\PassOptionsToPackage{numbers}{natbib}
\usepackage{natbib}
\usepackage[utf8]{inputenc} 
\usepackage[T1]{fontenc}    
\usepackage{hyperref}       
\usepackage{url}            
\usepackage{booktabs}       
\usepackage{amsfonts}       
\usepackage{nicefrac}       
\usepackage{microtype}      
\usepackage{xcolor}         
\usepackage{geometry}

\usepackage{microtype}
\usepackage{graphicx}
\usepackage[caption = false]{subfig}
\usepackage{float}
\title{Discovering the Hidden Vocabulary of DALLE-2}
\date{}

\author{%
  Giannis Daras and Alexandros G. Dimakis \\
  University of Texas at Austin.  \\
  \texttt{giannisdaras@utexas.edu},  \texttt{dimakis@austin.utexas.edu}
}

\begin{document}

\maketitle

\begin{abstract}
We discover that DALLE-2 seems to have a hidden vocabulary that can be used to generate images with absurd prompts. For example, it seems that \texttt{Apoploe vesrreaitais} means birds and \texttt{Contarra ccetnxniams luryca tanniounons} (sometimes) means bugs or pests. We find that these prompts are often consistent in isolation but also sometimes in combinations. We present our black-box method to discover words that seem random but have some correspondence to visual concepts. This creates important security and interpretability challenges. 
\end{abstract}

\begin{figure}[!htp]
\centering
\includegraphics[width=0.8\textwidth]{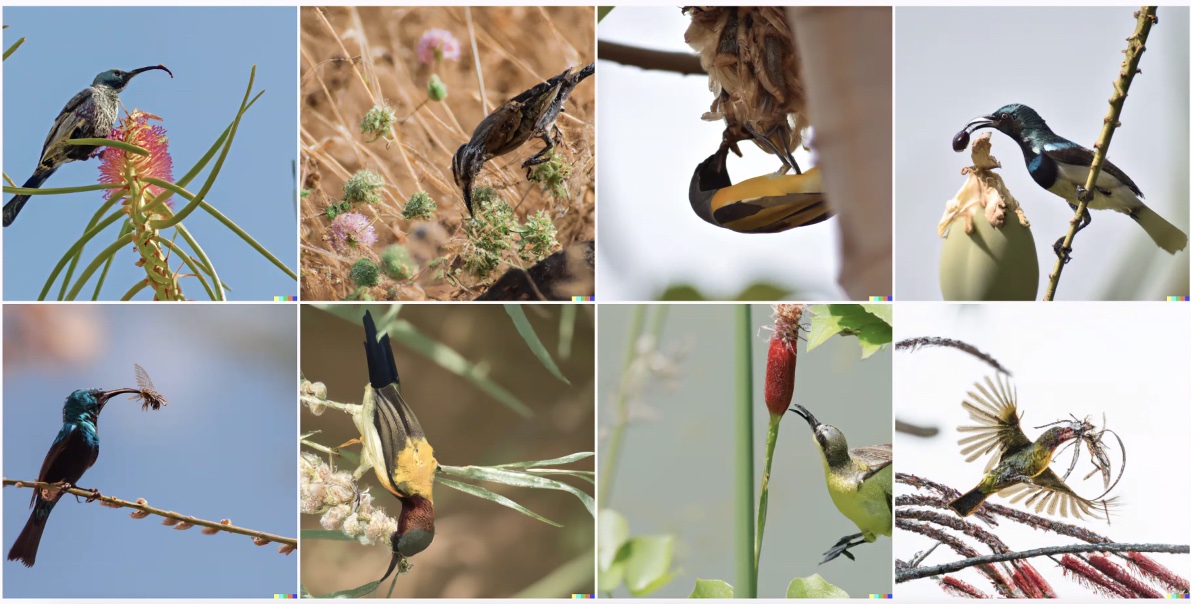}
\caption{Images generated with the prompt: ``\texttt{Apoploe vesrreaitais  eating Contarra ccetnxniams luryca tanniounons}''. We discover that DALLE-2 has its own vocabulary where \texttt{Apoploe vesrreaitais}  means birds and \texttt{Contarra ccetnxniams luryca tanniounons} (sometimes) means bugs. Hence, this prompt means ``Birds eating bugs''.}
\label{fig1}
\end{figure}

\section{Introduction}

DALLE \cite{dalle} and DALLE-2 \cite{dalle2} are deep generative models that take as input a text caption and generate images of stunning quality that match the given text. DALLE-2 uses Classifier-Free Diffusion Guidance~\citep{ho2021classifier} to generate high quality images. The conditioning is the CLIP~\cite{clip} text embeddings for the input text.


A known limitation of DALLE-2 is that it struggles with text. For example, text prompts such as: ``\texttt{An image of the word airplane}'' often lead to generated images that depict gibberish text.
We discover that this produced text is not random, but rather reveals a hidden vocabulary that the model seems to have developed internally. For example, when fed with this gibberish text, the model frequently produces airplanes. 

Some words from this hidden vocabulary can be learned and used to create absurd prompts that generate natural images. For example, it seems that \texttt{Apoploe vesrreaitais} means birds and \texttt{Contarra ccetnxniams luryca tanniounons} (sometimes) means bugs or pests.
We found that we can generate images of cartoon birds with prompts like \texttt{An image of a cartoon apoploe vesrreaitais} or even compose these terms to create birds eating bugs as shown in Figure \ref{fig1}.

\section{Discovering the DALLE-2 Vocabulary}
We provide a simple method to discover words of the DALLE-2 vocabulary. We use (in fact, we only have) query access to the model, through the API. We describe the method with an example. Assume that we want to find the meaning of the word: \texttt{vegetables}. Then, we can prompt DALLE-2 with one of the following sentences (or a variation of those):

\begin{itemize}
    \item ``\texttt{A book that has the word vegetables written on it.}''
    \item ``\texttt{Two people talking about vegetables, with subtitles.}''
    \item ``\texttt{The word vegetables written in $10$ languages.}''
\end{itemize}

\begin{figure}[!htp]
\centering 
\subfloat[Image generated with the prompt: ``\texttt{Two farmers talking about vegetables, with subtitles.}'']{\includegraphics[width=0.3\textwidth]{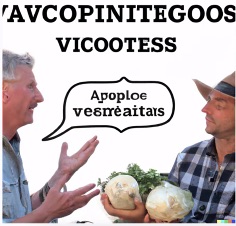}} \hspace{2mm}
\subfloat[Image generated with the prompt: ``\texttt{Vicootes.}'']{\includegraphics[width=0.3\textwidth]{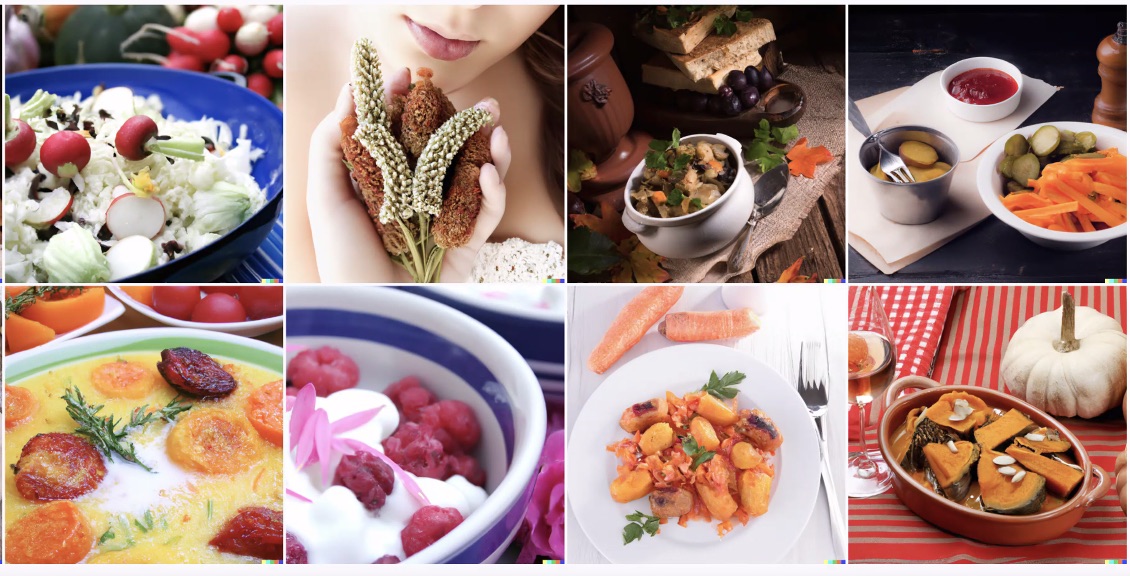}} \hspace{2mm}
\subfloat[Image generated with the prompt: ``\texttt{Apoploe vesrreaitais.}'']{\includegraphics[width =0.3\textwidth]{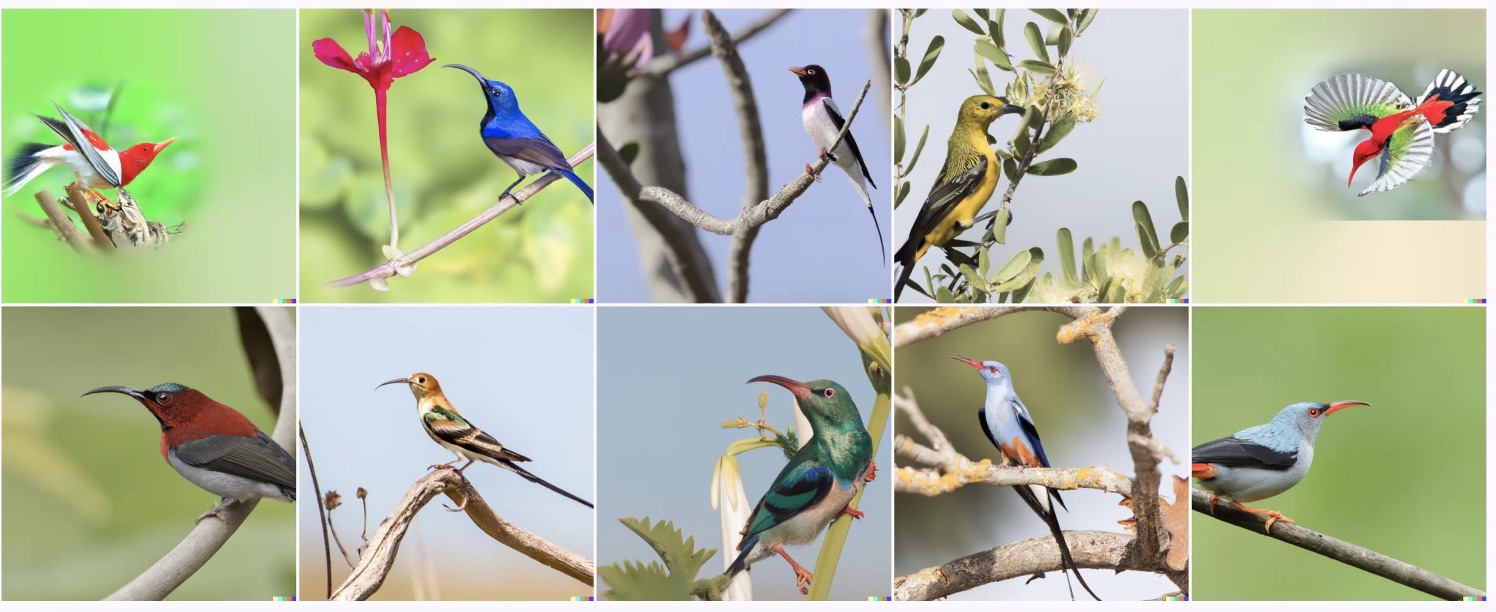}}
\caption{Illustration of our method for discovering words that seem random but can be understood by DALLE-2. We first query the model with the prompt: ``\texttt{Two farmers talking about vegetables, with subtitles.}''. The model generates an image with some gibberish text on it. We then prompt the model with words from this generated image, as shown in (b), (c). It seems that \texttt{Vicootes} means vegetables and \texttt{Apoploe vesrreaitais} means birds. Possibly farmers are talking about birds that interfere with their vegetables.}
\label{fig:farmers}
\end{figure}

For each of the above prompts, DALLE-2 usually creates images that have some text written text on it. The written text often seems gibberish to humans, as mentioned in the original DALLE-2 paper~\citep{dalle2} and also in the preliminary evaluation of the system by ~\citet{marcus2022preliminary}. However, we make the surprising observation that this text is not as random as it initially appears. In many cases, it is strongly correlated to the word we are looking to translate. For example, if we prompt DALLE-2 with the text: ``\texttt{Two farmers talking about vegetables, with subtitles.}'', we get the image shown in Figure \ref{fig:farmers}(a). We parse the text that appears in the images and we prompt the model with it as shown in Figure \ref{fig:farmers}(b), (c). It seems that \texttt{Vicootes} means vegetables and \texttt{Apoploe vesrreaitais} means birds. It appears that the farmers are talking about birds that interfere with their vegetables.

We note that this simple method doesn't always work. Sometimes, the generated text gives random images when prompted back to the model. However, we found that with some experimentation (selecting some words, running different produced texts, etc.) we can usually find words that appear random and are correlated with some visual concept (at least under some contexts). We encourage the interested readers to refer to the Limitations Section for more information.

\section{A Preliminary Study of the Discovered Vocabulary}
We do a very preliminary study of the properties of the found vocabulary of DALLE-2. 

\paragraph{Compositionality.} From the previous example, we learned that \texttt{Apoploe vesrreaitais} seems to mean birds. By repeating the experiment with the prompt about farmers, we also learn that: \texttt{Contarra ccetnxniams luryca tanniounons} may mean pests or bugs. An interesting question is whether we can compose these two concepts in a sentence, as we could do in an ordinary language. In Figure \ref{fig1}, we illustrate that this is possible, at least sometimes. The sentence: ``\texttt{Apoploe vesrreaitais eating Contarra ccetnxniams luryca tanniounons}'' gives images in which birds are eating bugs. We found that this happens for some, but not all of the generated images.

\begin{figure}[!htp]
    \centering
    \subfloat[Prompt: ``\texttt{Painting of Apoploe vesrreaitais}'']{\includegraphics[width=0.20\textwidth]{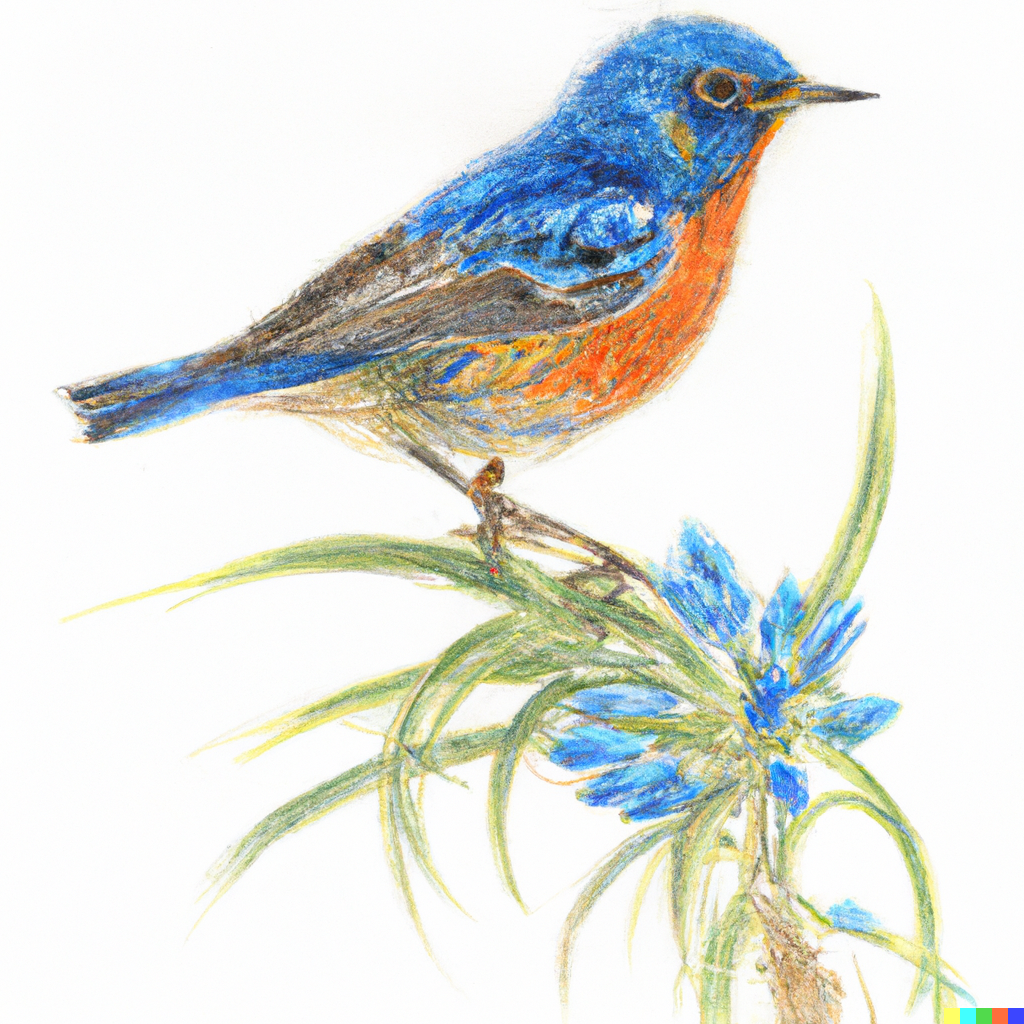}} \hspace{2mm}
    \subfloat[Prompt: ``\texttt{cartoon, Apoploe vesrreaitais}'']{\includegraphics[width=0.20\textwidth]{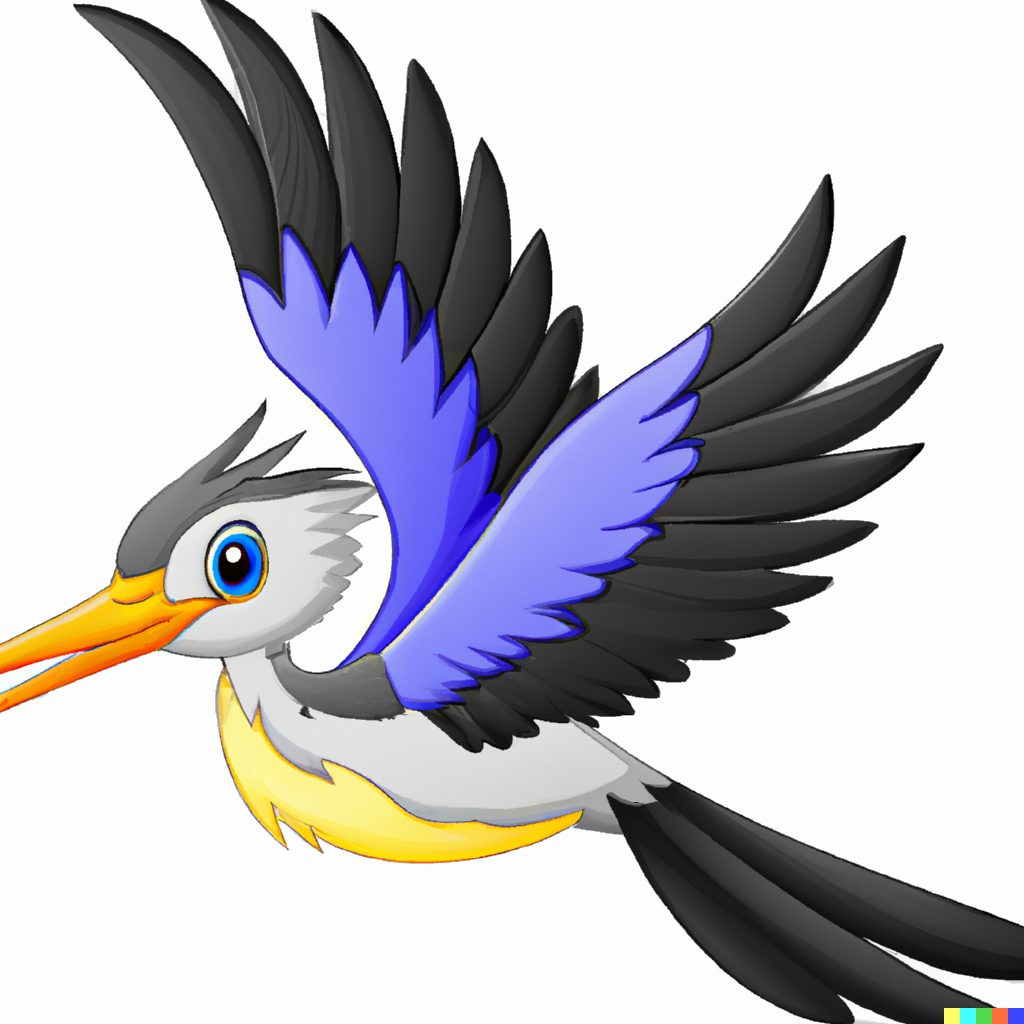}} \hspace{2mm}
        \subfloat[Prompt: ``\texttt{3-D rendering of Apoploe vesrreaitais}'']{\includegraphics[width=0.20\textwidth]{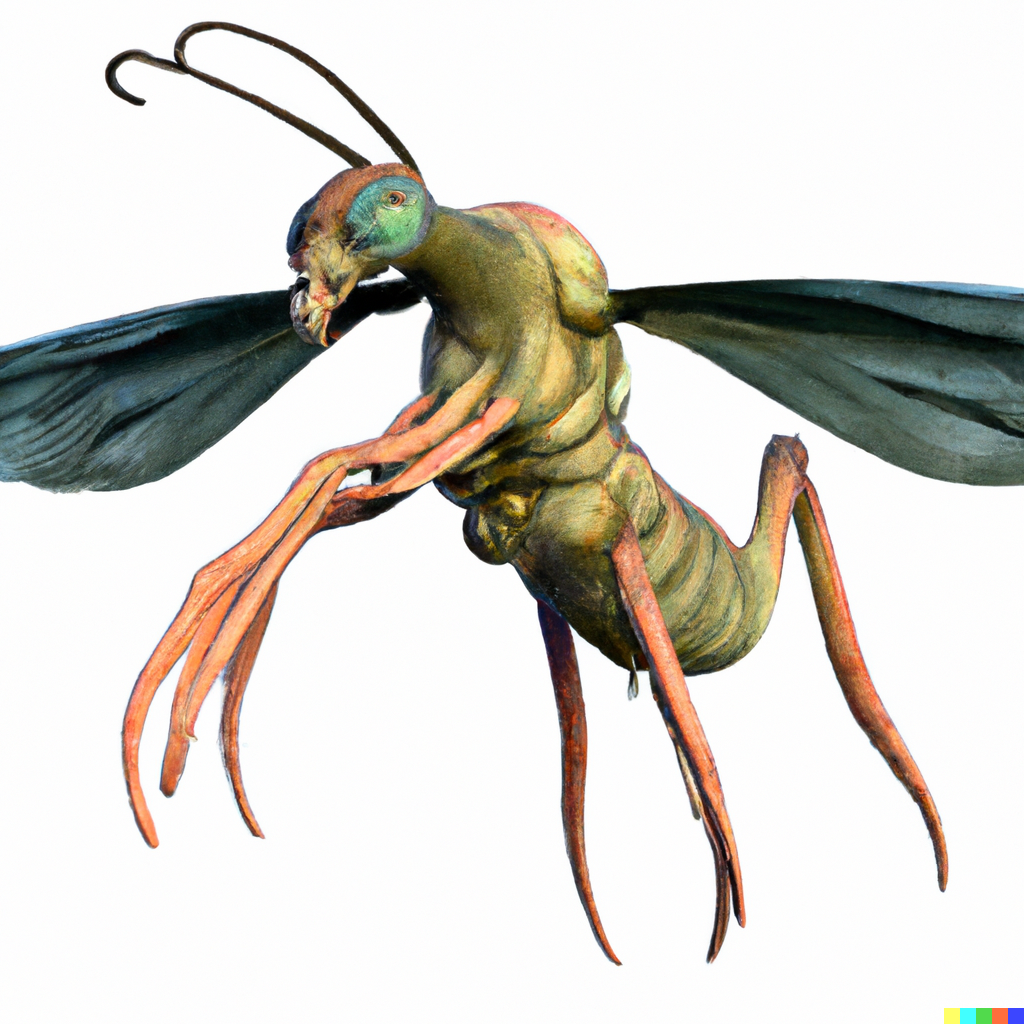}} \hspace{2mm} \subfloat[Prompt: ``\texttt{line art, Apoploe vesrreaitais}'']{\includegraphics[width=0.20\textwidth]{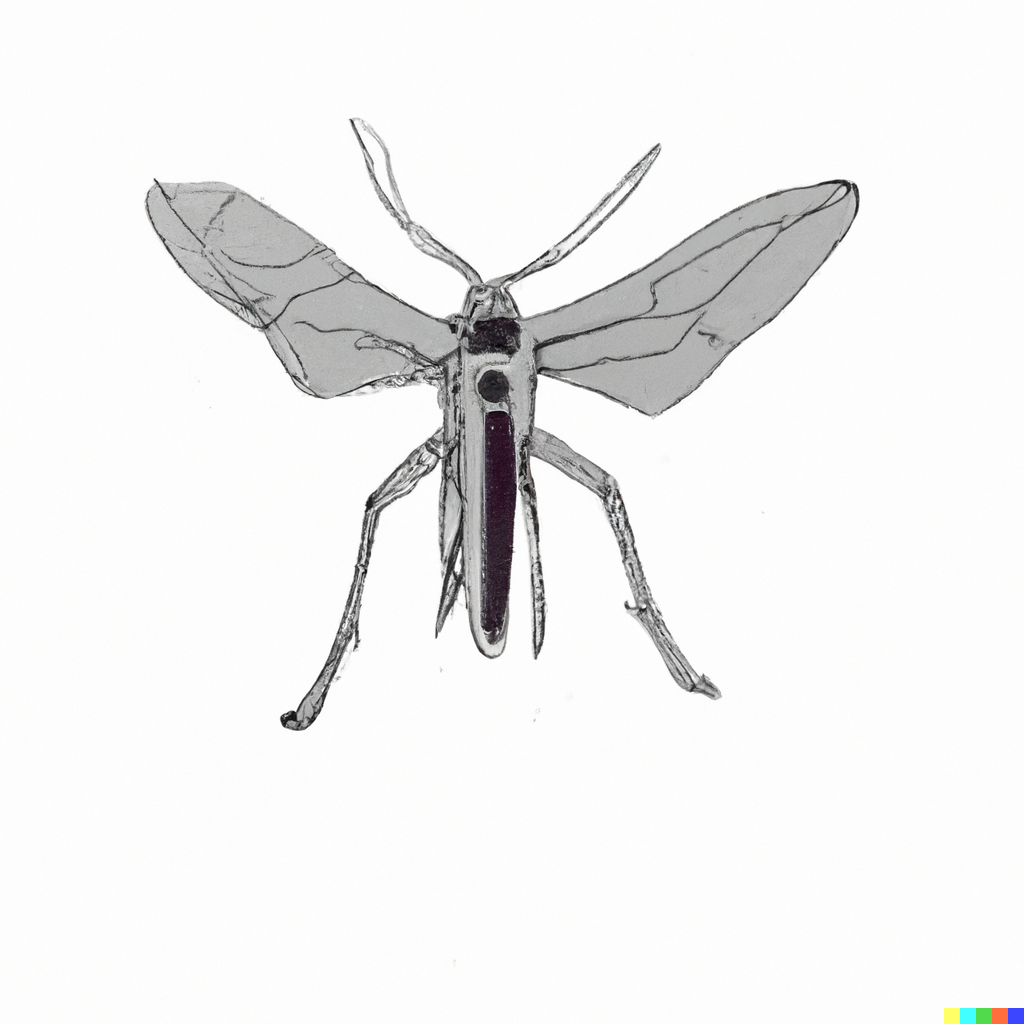}}
    \caption{Illustration of DALLE-2 generations for \texttt{Apoploe vesrreaitais} under different styles. The visual concept of ``something that flies'' is maintained across the different styles.}
    \label{fig:styles}
\end{figure}

\paragraph{Style Transfer.} DALLE-2 is capable of generating images of some concept under different styles that can be specified in the prompt~\citep{dalle2}. For example, one might ask for a photorealistic image of an apple or a line-art showing an apple. We test whether the discovered words, (e.g. \texttt{Apoploe vesrreaitais}) correspond to visual concepts that can be transformed into different styles, depending on the context of the prompt. The results of this experiment are shown in Figure \ref{fig:styles}. It seems that the prompt sometimes leads to flying insects as opposed to birds. 

\begin{figure}[!htp]
\begin{tabular}{@{}cc@{}}
    \raisebox{-\height}{\includegraphics[width=0.45\textwidth]{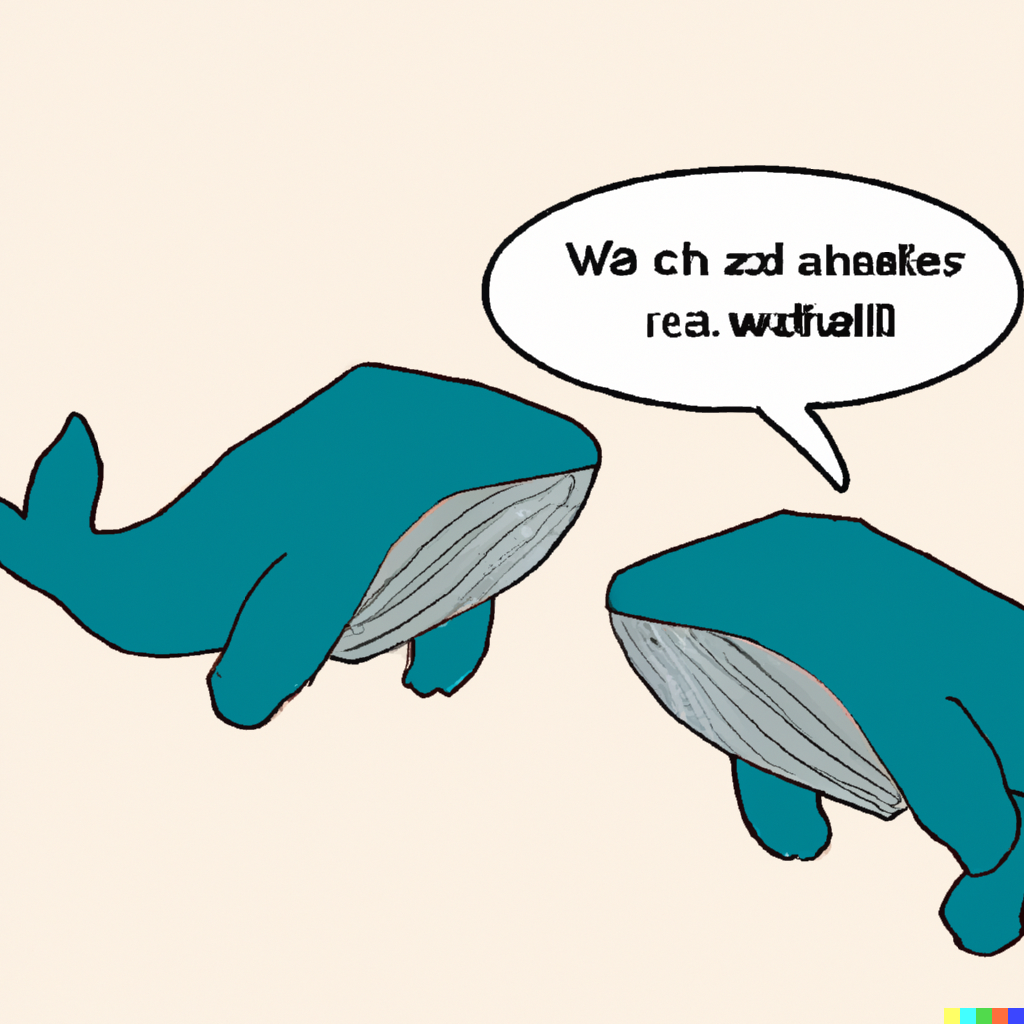}} & 
    \begin{tabular}[t]{@{}cc@{}}
        \raisebox{-\height}{\includegraphics[width=0.2\textwidth]{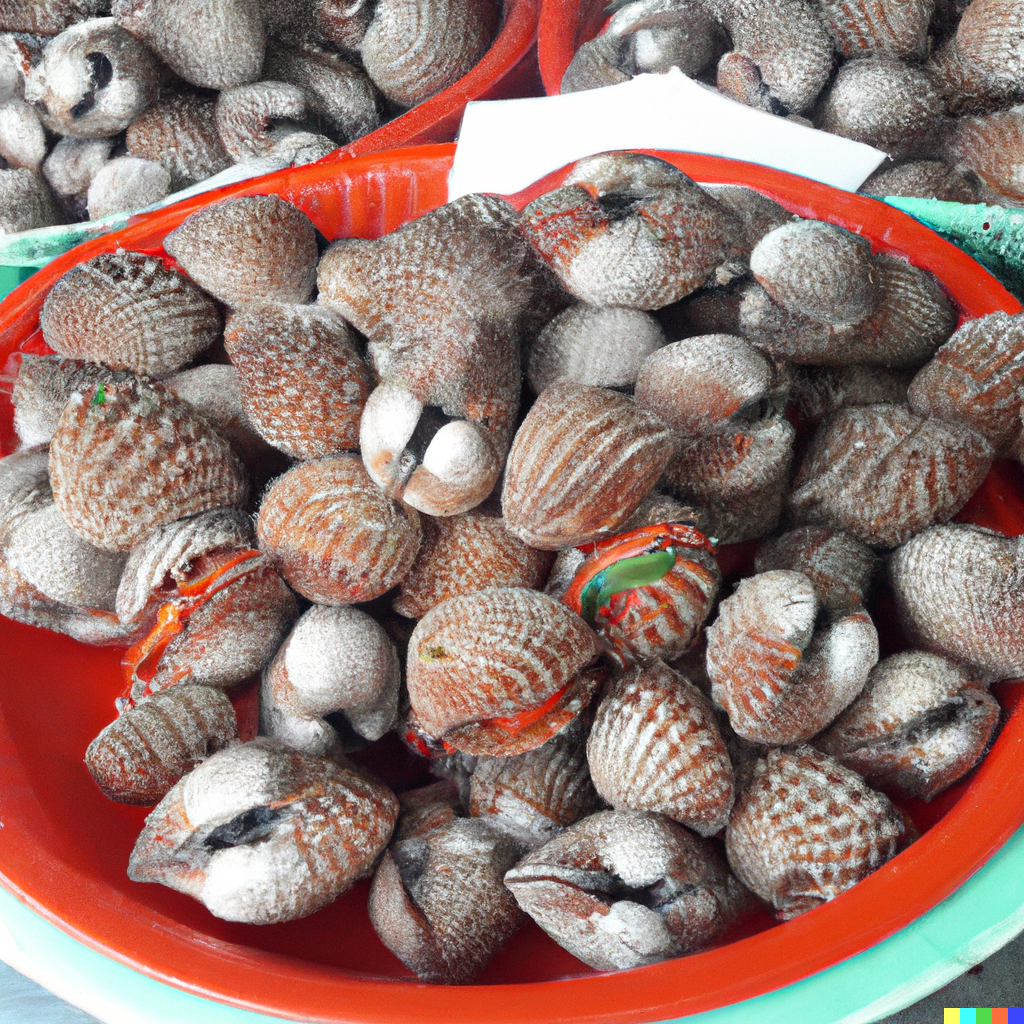}} & 
        \raisebox{-\height}{\includegraphics[width=0.2\textwidth]{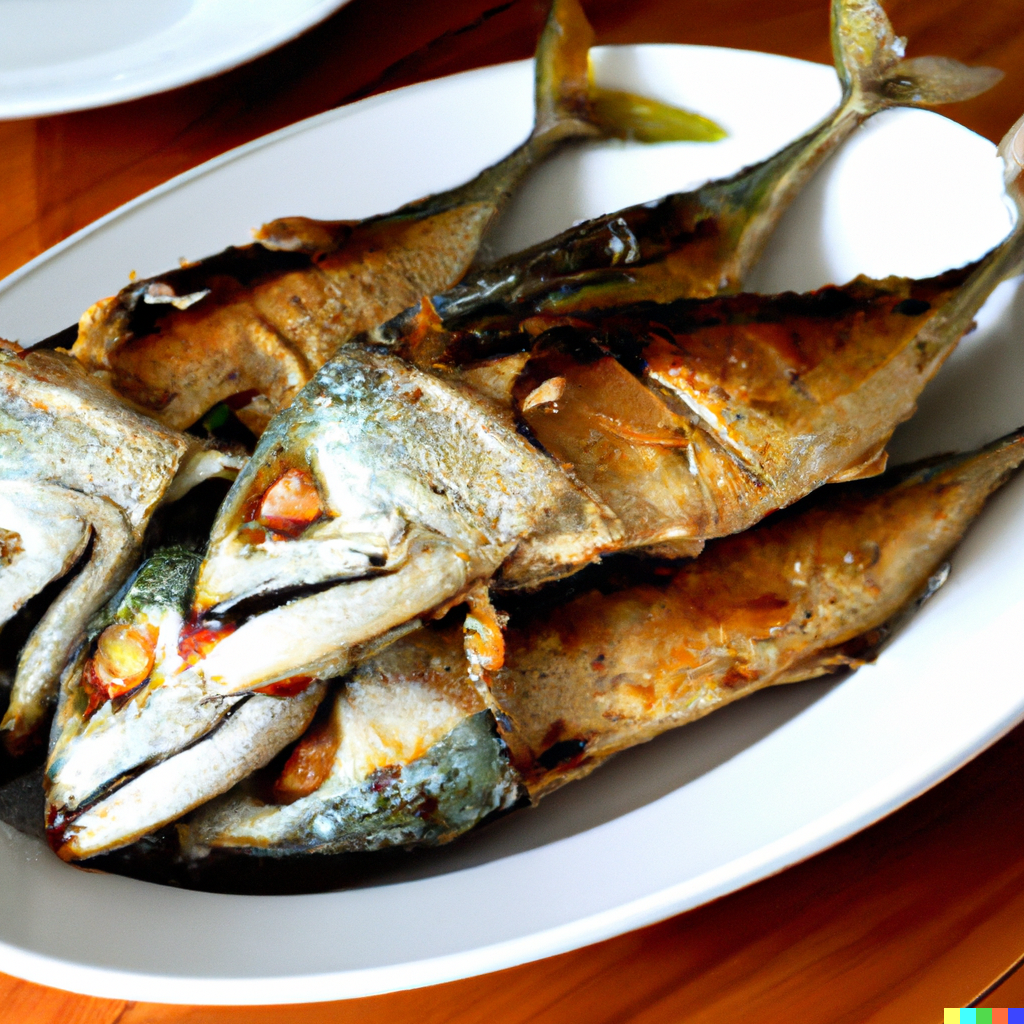}} \\[1.8cm]
        \raisebox{-\height}{\includegraphics[width=0.2\textwidth]{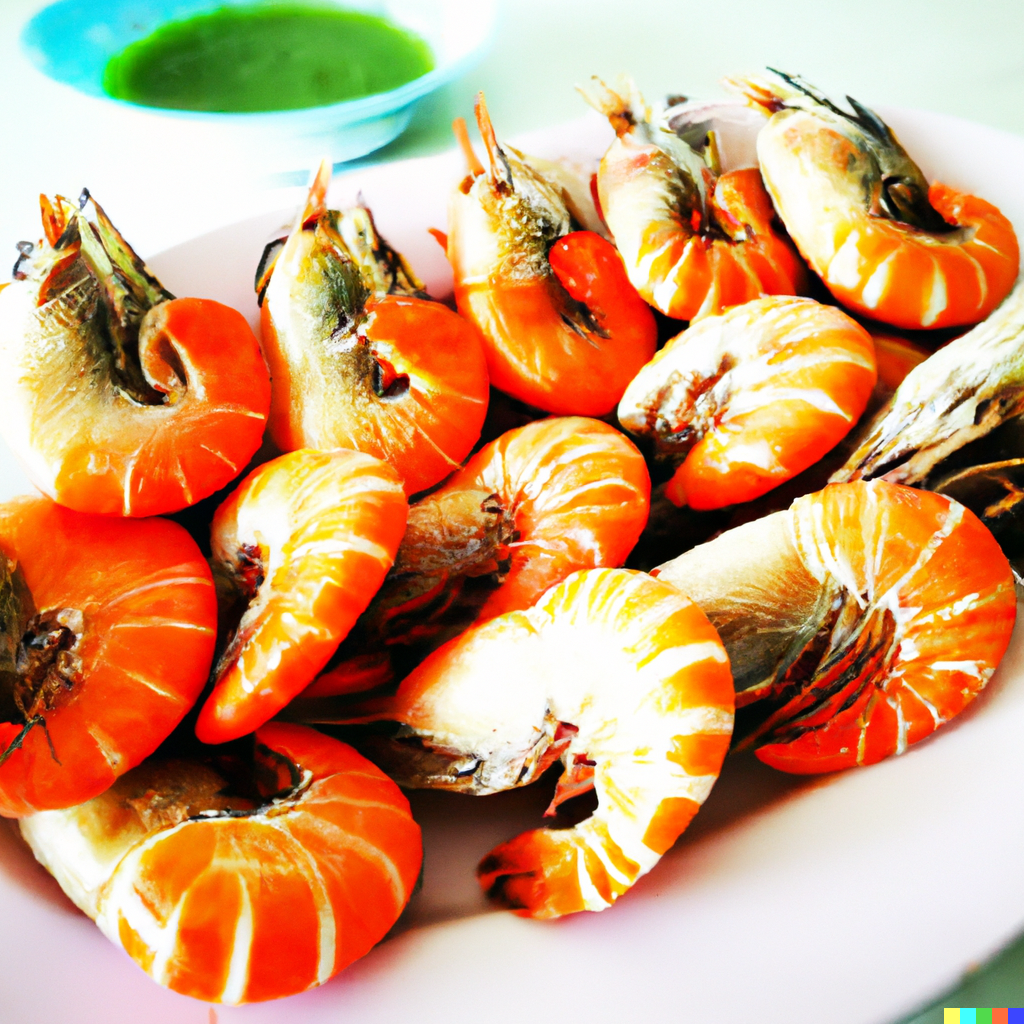}} & 
        \raisebox{-\height}{\includegraphics[width=0.2\textwidth]{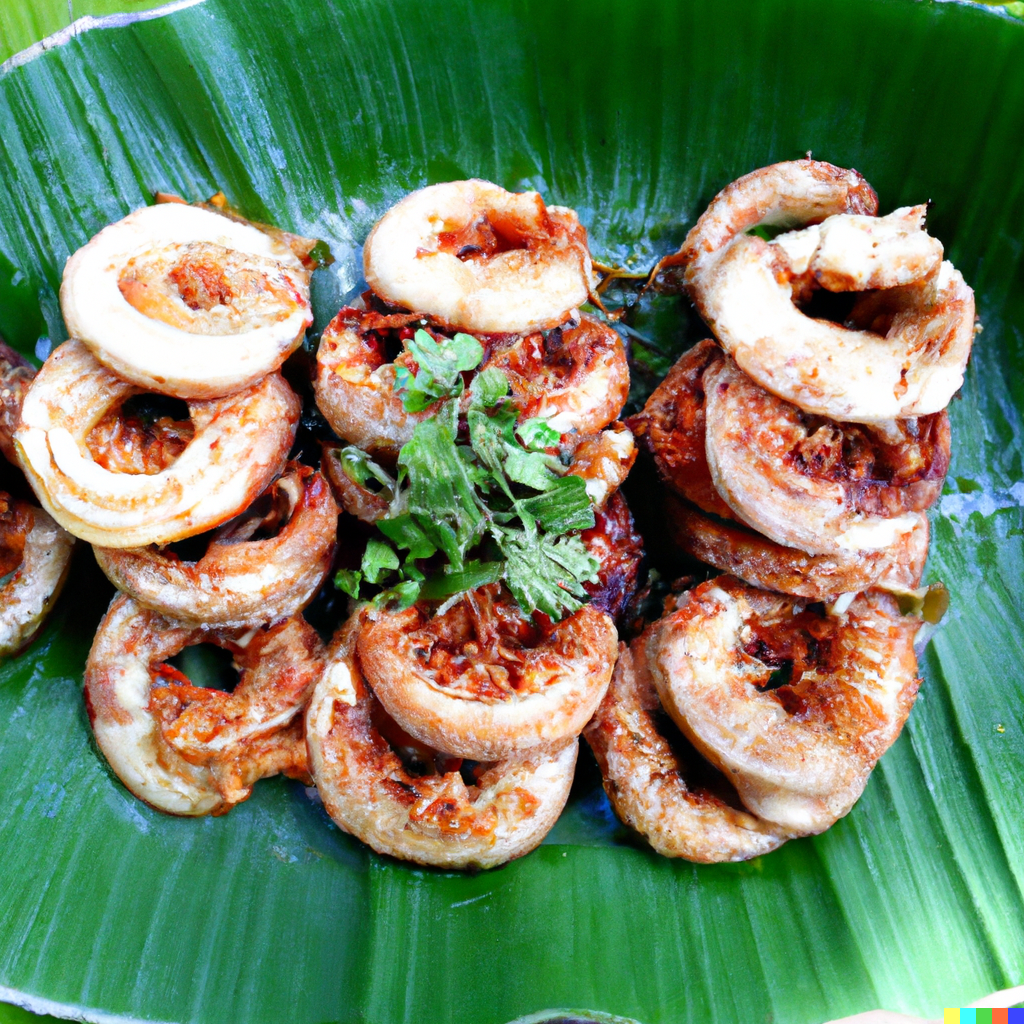}}
    \end{tabular}
\end{tabular}
    \label{fig:whales}
    \caption{Left: Image generated with the prompt: ``\texttt{Two whales talking about food, with subtitles.}''. Right: Images generated with the prompt: ``\texttt{Wa ch zod ahaakes rea.}''. The gibberish text, ``\texttt{Wa ch zod ahaakes rea.}'', produces images that are related to the caption and the visual output of the first image.}
\end{figure}

\paragraph{Text's consistency with the caption and the generated image.} Recall the example with the farmers. The prompt was: ``\texttt{Two farmers
talking about vegetables, with
subtitles.}''. From this example, we discovered the word vegetables, but also the word birds. It is very plausible that two farmers would be talking about birds and hence this opens the very interesting question of whether the text outputs of DALLE-2 are consistent with the text conditioning and the generated image. 
Our initial experiments show that sometimes we get gibberish text that translates to visual concepts that match the caption that created the gibberish text in the first place. For example, the prompt: ``\texttt{Two whales talking about food, with subtitles.}'' generates an image with the text ``\texttt{Wa ch zod ahaakes rea.}'' (or at least something close to that). We feed this text as prompt to the model and in the generated images we see seafood. This is shown in Figure \ref{fig:whales}. It seems that the gibberish text indeed has a meaning that is sometimes aligned with the text-conditioning that produced it.

\section{Security and Interpretability Challenges} 

There are many interesting directions for future research. It was not clear to us if some of the gibberish words are mispellings of normal words in different languages, but we could not find any such examples in our search. For many of the prompts, the origins of these words remains confusing and some of the words were not as consistent as others in our preliminary experiments. Another interesting question is if Imagen~\citep{imagen} has a similar hidden vocabulary given that it was trained with a language model as opposed to CLIP.
We conjecture that our prompts are adversarial examples for CLIP's~\citep{clip} 
text encoder, i.e. the vector representation of  \texttt{Apoploe vesrreaitais} is close to the representation of \texttt{bird}. It is natural to use other methods (e.g. white box) of adversarial attacks on CLIP to generate absurd text prompts that produce target images in DALLE2.

\paragraph{Robustness and Limitations.} One of the central questions is how consistent this method is. For example, our preliminary study shows that prompts like  
\texttt{Contarra ccetnxniams luryca tanniounons} sometimes produces bugs and pests (about half of the generated images) and sometimes different images, mostly animals. We found that 
\texttt{Apoploe vesrreaitais} is much more robust and can be combined in various ways as we show. We also want to emphasize that finding other robust prompts is challenging and requires a lot of experimentation. In our experiments we tried various ways of making DALLE generate images, selected parts of the generated text and tested its consistency. However, even if this method works for a few gibberish prompts (that are hard to find), this is still a big interpretability and security problem. If a system behaves in wildly unpredictable ways, even if this happens rarely and under unexpected conditions like gibberish prompts, this is still a significant concern, especially for some applications.

The first security issue relates to using these gibberish prompts as backdoor adversarial attacks or ways to circumvent filters.  Currently, Natural Language Processing systems filter text prompts that violate the policy rules and gibberish prompts may be used to bypass these filters. More importantly, absurd prompts that consistently generate images challenge our confidence in these big generative models. Clearly more foundational research is needed in understanding these phenomena and creating robust language and image generation models \textit{that behave as humans would expect}.

\section{Acknowledgements}
The authors would like to acknowledge the Institute for the Foundations of Machine Learning (IFML) and the National Science Foundation (NSF) for their generous support. We would like to thank Ludwig Schmidt, Rachael Tatman and others on Twitter who provided constructive feedback. We also thank OpenAI for providing access to their model through the API.

\bibliography{citations.bib}

\end{document}